\documentclass[10pt]{article}

\usepackage[T1]{fontenc}
\usepackage[utf8]{inputenc}
\usepackage{lmodern}
\usepackage[margin=1in]{geometry}

\usepackage{graphicx}
\usepackage{amsmath}
\usepackage{amssymb}
\usepackage{amsfonts}
\usepackage{booktabs}
\usepackage{multirow}
\usepackage{microtype}
\usepackage{xcolor}
\usepackage{url}
\usepackage{nicefrac}
\usepackage{placeins}

\usepackage{tikz}
\usetikzlibrary{arrows.meta, decorations.pathreplacing}

\usepackage{hyperref}
\hypersetup{
  colorlinks=true,
  linkcolor=blue!60!black,
  citecolor=green!40!black,
  urlcolor=blue!60!black,
}
\let\citep\cite
\let\citet\cite

\title{baseVultra: Compact Nonlinear Recurrence\\with Reallocated Capacity}

\author{Linke Zhang\thanks{This manuscript is a revised version based on peer review feedback from WAIC Academic 2026 (1 weak accept, 2 borderline).}\\
Southwest University\\
\texttt{zlk2270252364@email.swu.edu.cn}}

\date{August 2026}

\begin{document}
\maketitle

\begin{abstract}
Nonlinear recurrent networks remain useful for streaming sequence modeling, yet gated cell design is still dominated by LSTM and GRU. We study whether recurrent computation can be reallocated away from gate transforms and toward candidate generation. We propose baseVultra, a compact recurrent cell that concentrates the dense recurrent budget on a single $H \times H$ candidate-generation matrix, while replacing recurrent gate matrices with $O(H)$ diagonal interactions. The candidate update combines this dense branch with two low-rank auxiliary branches, an EMA cache, and learnable output blending; the gates additionally receive lightweight second-order access to $\mathbf{h}_{t-2}$ through diagonal terms. On TinyShakespeare character-level language modeling, baseVultra achieves a validation loss of $1.594$ in the compact one-layer 200K setting, compared with $1.604$ for a parameter-matched LSTM and $1.616$ for a parameter-matched GRU, while its dense recurrent matrix-vector computation is about $54\%$ of GRU under matched hidden size for this configuration. On text8, a roughly $45\times$ larger benchmark, baseVultra is competitive with full-gate cells ($1.854$ bpc vs.\ $1.849$ for GRU and $1.837$ for LSTM). On formal-language probes, the largest gains appear under constrained hidden-state capacity ($0.991$ vs.\ $0.761$ GRU and $0.691$ LSTM OOD2 accuracy on Counter at $H=16$) and vanish as hidden size grows. These results characterize where and when candidate-focused recurrent capacity allocation helps: the benefit is concentrated in compact regimes, while full-gate cells remain competitive at scale.

\end{abstract}

\section{Introduction}
\label{sec:intro}
LSTM and GRU have long served as practical baselines for nonlinear recurrent neural networks. Their success established a common design pattern for gated recurrent cells: use several full-rank recurrent transformations for gate computation and candidate generation, and update the hidden state primarily from the previous hidden state. Although recent sequence modeling research has shifted toward Transformers and recurrent-like linear-time architectures, nonlinear recurrent models remain attractive in settings where constant-size state, streaming inference, and temporal inductive bias are important.

Despite these advantages, practical nonlinear RNN design beyond LSTM and GRU has seen relatively limited adoption. We argue that this reflects not necessarily an inherent limitation of nonlinear recurrence, but a set of architectural assumptions that remain worth revisiting. In particular, conventional gated RNNs allocate multiple full-rank $H \times H$ recurrent matrices to gates as well as candidate generation. In compact regimes, this may spend substantial recurrent capacity on gate computation rather than on constructing the candidate state. In addition, most mainstream gated cells are effectively first-order in their hidden-state transition, conditioning updates mainly on $\mathbf{h}_{t-1}$ and leaving lightweight higher-order dependence underexplored.

This paper studies a simple alternative: concentrate the dense recurrent budget on candidate generation and make the recurrent gates lightweight. We propose baseVultra, a compact nonlinear recurrent cell based on recurrent capacity reallocation. baseVultra uses a single dense $H \times H$ recurrent matrix for the main candidate branch, while replacing full recurrent gate matrices with $O(H)$ diagonal interactions. Under matched hidden size, this reduces dense recurrent matrix-vector computation to about $54\%$ of GRU for the compact configuration used in our experiments.\footnote{This ratio is configuration-dependent; see Sec.~\ref{sec:method} for the general expression.} The goal is not to remove gating, but to allocate the expensive full-rank recurrent computation to the part of the transition that directly constructs the candidate state.

As a secondary design element, baseVultra introduces higher-order temporal access without adding another dense recurrent matrix: $\mathbf{h}_{t-2}$ enters the gating dynamics through element-wise diagonal terms, yielding second-order dependence in the gate-conditioned transition. We stress that this is a minor component: our ablations (Sec.~\ref{sec:ablation}) and targeted probes (Sec.~\ref{sec:formal}) show its effect to be marginal in most settings, and we report it as a capacity-dependent refinement rather than a headline contribution. This does not imply that first-order recurrent models are theoretically unable to represent two-step dependencies; a sufficiently large hidden state can encode earlier history. Rather, direct access to $\mathbf{h}_{t-2}$ may modestly reduce the burden on $\mathbf{h}_{t-1}$ when hidden-state capacity is constrained.

baseVultra further enriches candidate generation through a differentiated multi-branch design. A full-rank main branch operates in parallel with two low-rank auxiliary branches. The update gate $\mathbf{z}_t$ amplifies the main branch through $(1+\tilde{\boldsymbol{\beta}} \odot \mathbf{z}_t)$ and suppresses the primary low-rank branch through $(1-\mathbf{z}_t)$, while a separate gate $\mathbf{c}_t$ independently controls the secondary low-rank branch. The cell also includes a lightweight EMA cache and learnable output blending, providing additional temporal smoothing and output control at modest cost.

We evaluate baseVultra on character-level language modeling at two scales (TinyShakespeare and text8) and on formal-language tasks designed to probe structured sequence behavior. On TinyShakespeare, baseVultra obtains a validation loss of $1.594$ in the compact one-layer setting, compared with $1.616$ for a parameter-matched GRU and $1.604$ for a parameter-matched LSTM. On the roughly $45\times$ larger text8 benchmark, baseVultra is competitive with full-gate cells but does not outperform them, marking the operating boundary of the method. On low-capacity formal-language probes, baseVultra shows the clearest advantage on counting and delayed-copy tasks, while this advantage disappears as hidden size increases. Throughout, we scope our claims to the compact regime where the benefit is concentrated.

\paragraph{Contributions.}
Our contributions are summarized as follows:
\begin{itemize}
\item We propose baseVultra, a compact nonlinear recurrent cell that concentrates the dense $H \times H$ recurrent matrix on candidate generation while reducing recurrent gate computation to $O(H)$ diagonal interactions, with two low-rank auxiliary branches, a fixed global gate-scaling constant $\alpha$, an element-wise adaptive gain vector $\boldsymbol{\beta}$, learnable output blending, and a lightweight second-order gating term.
\item We characterize where the reallocation strategy helps: through parameter-matched experiments on TinyShakespeare and text8, together with a capacity sweep over formal-language probes, we show that the advantage is concentrated in compact regimes and vanishes as hidden-state capacity grows.
\item We isolate the contribution of each design component through a component-wise ablation and a minimal-core baseline with identical parameters, showing that the differentiated multi-branch candidate design is the primary driver of the gains, while the second-order term and the cache are minor refinements.
\item We measure practical efficiency directly on two GPUs (wall-clock time and peak memory), and scope the efficiency claim to dense recurrent matrix-vector FLOPs and parameter count under matched hidden size.
\item We use a 7-task formal-language suite as structural probes of recurrent behavior, covering counting, state locking, copying, delayed dependency, and hierarchical matching.
\end{itemize}
\FloatBarrier

\section{Method}
\label{sec:method}

\begin{figure}[t]
\centering
\resizebox{\linewidth}{!}{%
\begin{tikzpicture}[
  box/.style={rectangle, draw, rounded corners=2pt,
              minimum width=2.4cm, minimum height=0.65cm,
              font=\small, align=center},
  dbox/.style={rectangle, draw, rounded corners=2pt,
               minimum width=2.4cm, minimum height=0.65cm,
               font=\small, align=center, fill=gray!12,
               densely dashed},
  lbox/.style={rectangle, draw, rounded corners=2pt,
               minimum width=2.4cm, minimum height=0.5cm,
               font=\footnotesize, align=center, fill=blue!8},
  arr/.style={->, >=Stealth, thick},
  darr/.style={->, >=Stealth, thick, densely dashed, gray},
]

\node[font=\normalsize\bfseries] at (0, 3.0) {GRU};

\node[box, fill=orange!15] (gr) at (0, 1.9)
  {Reset gate\\$W_r \mathbf{h}_{t-1}$\\{\tiny $H\!\times\!H$}};

\node[box, fill=orange!15] (gz) at (0, 0.6)
  {Update gate\\$W_z \mathbf{h}_{t-1}$\\{\tiny $H\!\times\!H$}};

\node[box, fill=orange!25] (gc) at (0,-0.7)
  {Candidate\\$W_n(\mathbf{r}_t\!\odot\!\mathbf{h}_{t-1})$\\
   {\tiny $H\!\times\!H$}};

\node[font=\small] (lh) at (-2.4, 0.6) {$\mathbf{h}_{t-1}$};
\draw[thick, orange!60] (lh.east) -- (-1.5, 0.6);
\draw[thick, orange!60] (-1.5, 1.9) -- (-1.5, -0.7);
\draw[arr, orange!60] (-1.5, 1.9)  -- (gr.west);
\draw[arr, orange!60] (-1.5, 0.6)  -- (gz.west);
\draw[arr, orange!60] (-1.5, -0.7) -- (gc.west);

\draw[decorate,
      decoration={brace, amplitude=6pt, mirror},
      thick, orange!60!black]
  (1.45,-1.15) -- (1.45, 2.35)
  node[midway, right=8pt, font=\small,
       text=orange!70!black, align=left]
  {$3\!\times$ full-rank\\recurrent\\matrices};

\node[font=\small, align=center, text=orange!70!black]
  at (0, -1.8) {Dense cost: $3H^2$};

\draw[very thick, gray!35] (3.6,-2.2) -- (3.6, 3.2);

\node[font=\normalsize\bfseries] at (7.2, 3.0) {baseVultra};

\node[box, fill=green!18] (bv_cand) at (7.2, 2.0)
  {Candidate\\$W_{vh}(\mathbf{r}_t\!\odot\!\mathbf{h}_{t-1})$\\
   {\tiny \textbf{single} $H\!\times\!H$}};

\node[dbox] (bv_gate) at (7.2, 0.6)
  {Gate-conditioned transition\\
   $\mathbf{r}_t,\mathbf{z}_t,\mathbf{c}_t$\\
   {\tiny $O(H)$ diagonal}};

\node[lbox] (bv_lr1) at (7.2,-0.45)
  {\footnotesize Low-rank $r_1$};
\node[lbox] (bv_lr2) at (7.2,-1.05)
  {\footnotesize Low-rank $r_2$};

\node[font=\small] (rh) at (4.2, 0.6) {$\mathbf{h}_{t-1}$};
\draw[thick] (rh.east) -- (4.8, 0.6);
\draw[thick] (4.8, 2.0) -- (4.8, -1.05);
\draw[arr] (4.8, 2.0)  -- (bv_cand.west);
\draw[arr] (4.8, 0.6)  -- (bv_gate.west);
\draw[arr] (4.8, -0.45) -- (bv_lr1.west);
\draw[arr] (4.8, -1.05) -- (bv_lr2.west);

\node[font=\small] (ht2) at (10.2, 0.6) {$\mathbf{h}_{t-2}$};
\draw[darr] (ht2.west) -- (bv_gate.east)
  node[midway, above, font=\scriptsize, text=gray] {diag};

\draw[decorate,
      decoration={brace, amplitude=6pt},
      thick, green!45!black]
  (10.5, 2.40) -- (10.5,-1.15)
  node[midway, right=10pt, font=\small,
       text=green!45!black, align=left]
  {$1\!\times$ full-rank\\$+$ diagonal gates\\$+$ low-rank paths};

\node[font=\small, align=center, text=green!45!black]
  at (7.2,-1.95)
  {Dense cost: $H^2\!+\!2H(r_1\!+\!r_2\!+\!r_c)$\\
   {\small $\approx 54\%$ of GRU (compact setting)}};

\end{tikzpicture}
}%
\caption{%
\textbf{Recurrent capacity reallocation in baseVultra.}
Conventional GRU uses three full-rank $H \times H$ recurrent
matrices for hidden-state updates.
baseVultra concentrates the single dense $H \times H$ matrix
on candidate generation and replaces recurrent gate transforms
with $O(H)$ diagonal interactions, two low-rank auxiliary
branches, and explicit second-order access to $\mathbf{h}_{t-2}$
without adding another dense recurrent matrix.
For the compact configuration $(H=420,\,r_1=64,\,r_2=r_c=32)$,
this reduces dense recurrent matrix-vector cost to
approximately $54\%$ of GRU under matched hidden size.}
\label{fig:arch}
\end{figure}
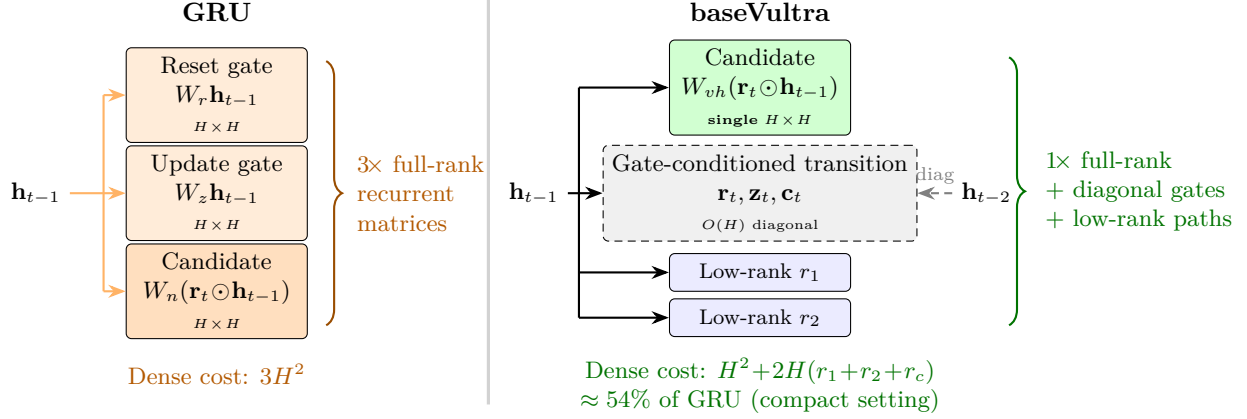

We propose \textbf{baseVultra}, a compact nonlinear recurrent cell built around recurrent capacity reallocation. The cell uses a single dense $H \times H$ recurrent matrix for candidate generation, while recurrent gate-history interactions are reduced to element-wise diagonal terms. In addition, baseVultra introduces second-order dependence in the gate-conditioned transition through $\mathbf{h}_{t-2}$, enriches candidate construction with two low-rank auxiliary branches, and augments the recurrent update with a lightweight exponential-memory cache and learnable output blending.

Let $\mathbf{x}_t \in \mathbb{R}^{d_{\mathrm{in}}}$ denote the input at time step $t$. The recurrent state consists of the previous output state $\mathbf{h}_{t-1} \in \mathbb{R}^{H}$, the second-order history state $\mathbf{h}_{t-2} \in \mathbb{R}^{H}$, and an auxiliary cache state $\mathbf{m}^{\mathrm{cache}}_{t-1} \in \mathbb{R}^{H}$. We use boldface symbols for vectors and reserve $\alpha \in \mathbb{R}$ as a fixed global gate-scaling constant, set to $\sqrt{e/2}\approx1.1658$ in all experiments and not tuned per task.

\paragraph{Input projection.}
We first project the input into four $H$-dimensional components:
\begin{equation}
\mathbf{ih}_t = W_x \mathbf{x}_t + \mathbf{b}_x \in \mathbb{R}^{4H},
\end{equation}
and split it as
\begin{equation}
\mathbf{ih}_t =
\left[
\mathbf{ih}_t^{(r)};
\mathbf{ih}_t^{(z)};
\mathbf{ih}_t^{(c)};
\mathbf{ih}_t^{(v)}
\right].
\end{equation}

\paragraph{Second-order diagonal gating.}
The three gates depend on both first-order and second-order hidden history through element-wise diagonal interactions:
\begin{align}
\mathbf{r}_t &= \sigma\!\left(
\alpha \bigl(
\mathbf{ih}_t^{(r)}
+ \mathbf{w}_{rh} \odot \mathbf{h}_{t-1}
+ \mathbf{w}_{rh2} \odot \mathbf{h}_{t-2}
+ \mathbf{b}_r
\bigr)\right), \\
\mathbf{z}_t &= \sigma\!\left(
\alpha \bigl(
\mathbf{ih}_t^{(z)}
+ \mathbf{w}_{zh} \odot \mathbf{h}_{t-1}
+ \mathbf{w}_{zh2} \odot \mathbf{h}_{t-2}
+ \mathbf{b}_z
\bigr)\right), \\
\mathbf{c}_t &= \sigma\!\left(
\alpha \bigl(
\mathbf{ih}_t^{(c)}
+ \mathbf{w}_{ch} \odot \mathbf{h}_{t-1}
+ \mathbf{w}_{ch2} \odot \mathbf{h}_{t-2}
+ \mathbf{b}_c
\bigr)\right).
\end{align}

\paragraph{Dense main branch.}
The only dense recurrent $H \times H$ matrix is allocated to the main candidate-generation path:
\begin{equation}
\mathbf{m}_t = W_{vh}(\mathbf{r}_t \odot \mathbf{h}_{t-1}).
\end{equation}

\paragraph{Low-rank auxiliary branches.}
Two low-rank branches provide additional corrective pathways:
\begin{align}
\mathbf{u}^{(1)}_t &= U_1 (V_1 \mathbf{h}_{t-1}),
\qquad
V_1 \in \mathbb{R}^{r_1 \times H},
\quad
U_1 \in \mathbb{R}^{H \times r_1}, \\
\mathbf{u}^{(2)}_t &= U_2 (V_2 \mathbf{h}_{t-1}),
\qquad
V_2 \in \mathbb{R}^{r_2 \times H},
\quad
U_2 \in \mathbb{R}^{H \times r_2}.
\end{align}

\paragraph{Differentiated candidate fusion.}
We clip the adaptive gain vector element-wise:
\begin{equation}
\tilde{\boldsymbol{\beta}} = \mathrm{clip}(\boldsymbol{\beta}, 0, \beta_{\max}),
\end{equation}
and construct the candidate state as
\begin{equation}
\mathbf{v}_t =
\tanh\!\Big(
\mathbf{ih}_t^{(v)}
+ (1 + \tilde{\boldsymbol{\beta}} \odot \mathbf{z}_t) \odot \mathbf{m}_t
+ (1 - \mathbf{z}_t) \odot \mathbf{u}^{(1)}_t
+ (1 - \mathbf{c}_t) \odot \mathbf{u}^{(2)}_t
+ \Delta \mathbf{b}_v
\Big).
\end{equation}
Here, $\mathbf{z}_t$ simultaneously amplifies the dense main branch through $(1+\tilde{\boldsymbol{\beta}} \odot \mathbf{z}_t)$ and suppresses the primary low-rank branch through $(1-\mathbf{z}_t)$, while $\mathbf{c}_t$ independently controls the secondary low-rank branch.

\paragraph{State update.}
The gate-conditioned hidden update is
\begin{equation}
\hat{\mathbf{h}}_t
=
\mathbf{z}_t \odot \mathbf{h}_{t-1}
+
(1 - \mathbf{z}_t) \odot \mathbf{v}_t.
\end{equation}

\paragraph{EMA cache.}
To provide a soft long-range memory pathway, we introduce a low-rank projected EMA cache:
\begin{align}
\boldsymbol{\rho} &= \sigma(\boldsymbol{\rho}_{\mathrm{raw}}), \\
\mathbf{p}_t &= P_u(P_v \hat{\mathbf{h}}_t),
\qquad
P_v \in \mathbb{R}^{r_c \times H},
\quad
P_u \in \mathbb{R}^{H \times r_c}, \\
\mathbf{m}^{\mathrm{cache}}_t
&=
\boldsymbol{\rho} \odot \mathbf{m}^{\mathrm{cache}}_{t-1}
+
\mathbf{p}_t.
\end{align}

\paragraph{Output blending.}
The cache is injected into the hidden state before the final output projection:
\begin{align}
\hat{\mathbf{h}}^{\,c}_t
&=
\hat{\mathbf{h}}_t + \boldsymbol{\gamma} \odot \mathbf{m}^{\mathrm{cache}}_t, \\
\mathbf{h}_t
&=
\tanh\!\left(
\boldsymbol{\ell}_1 \odot \hat{\mathbf{h}}^{\,c}_t
+
\boldsymbol{\ell}_2 \odot \mathbf{h}_{t-1}
\right),
\end{align}
where $\boldsymbol{\ell}_1, \boldsymbol{\ell}_2 \in \mathbb{R}^{H}$ are learnable element-wise output blending coefficients. Before the first recurrent step, $\mathbf{h}_{t-1}$ is initialized to $\mathbf{h}_{\mathrm{init}}$, while $\mathbf{h}_{t-2}$ and the cache are initialized to zero.

\paragraph{State advancement.}
After each step, the recurrent history is shifted as
\begin{equation}
\mathbf{h}_{t-2} \leftarrow \mathbf{h}_{t-1},
\qquad
\mathbf{h}_{t-1} \leftarrow \mathbf{h}_{t}.
\end{equation}

\paragraph{Second-order gate-conditioned dependence.}
We do \emph{not} claim that the full output process $\mathbf{h}_t$ is strictly second-order Markov, because the EMA cache introduces an auxiliary memory pathway that carries longer-range information. Instead, our claim concerns the \emph{gate-conditioned recurrent transition}. Conditioned on the current input $\mathbf{x}_t$, the gates $(\mathbf{r}_t, \mathbf{z}_t, \mathbf{c}_t)$ and the candidate state $\mathbf{v}_t$ depend explicitly on $(\mathbf{h}_{t-1}, \mathbf{h}_{t-2})$ and are conditionally independent of $\mathbf{h}_{<t-2}$. In this sense, baseVultra introduces \emph{second-order gate-conditioned dependence}, while the EMA cache provides a separate auxiliary long-range memory mechanism. We measure the empirical contribution of this mechanism in Sec.~\ref{sec:ablation} and Sec.~\ref{sec:formal}; it is marginal in most settings.

Formally, let
$\mathcal{G}_t = \{\mathbf{r}_t, \mathbf{z}_t, \mathbf{c}_t, \mathbf{v}_t\}$
denote the set of gate-conditioned transition variables at time step $t$.
We define second-order gate-conditioned dependence as
\begin{equation}
p(\mathcal{G}_t \mid \mathbf{x}_{\le t}, \mathbf{h}_{<t})
=
p(\mathcal{G}_t \mid \mathbf{x}_t, \mathbf{h}_{t-1}, \mathbf{h}_{t-2}).
\end{equation}
The EMA cache is intentionally excluded from this definition, as it acts as an auxiliary memory state rather than part of the gate-conditioned transition itself.

\paragraph{Dense recurrent computation.}
Under matched hidden size, we compare baseVultra and GRU in terms of \emph{dense recurrent matrix-vector operations} per time step. For baseVultra, the recurrent hidden-state matrix-vector cost consists of one dense $H \times H$ main branch and three low-rank auxiliary projections:
\begin{equation}
C_{\mathrm{bV}}
=
H^2 + 2H(r_1 + r_2 + r_c).
\end{equation}
For a standard GRU, the corresponding recurrent cost is
\begin{equation}
C_{\mathrm{GRU}} = 3H^2.
\end{equation}
Therefore, the relative recurrent compute ratio is
\begin{equation}
\mathrm{ratio}
=
\frac{H^2 + 2H(r_1 + r_2 + r_c)}{3H^2}.
\end{equation}
With the compact configuration used in our experiments $(H=420,\; r_1=64,\; r_2=32,\; r_c=32)$, this gives $\mathrm{ratio} \approx 0.537$, i.e., approximately $54\%$ of the dense recurrent matrix-vector cost of a GRU under matched hidden size. This ratio is configuration-dependent: with fixed ranks it decreases toward $1/3$ as $H$ increases, whereas it remains approximately constant when the ranks scale proportionally with $H$.

\section{Experiments}
\label{sec:experiments}

\subsection{Experimental Setup}
\label{sec:setup}

\paragraph{Character-level language modeling.}
We evaluate baseVultra on TinyShakespeare, using the standard character-level setup with 65 unique characters, 1{,}003{,}854 training tokens, and 111{,}540 validation tokens. We report validation cross-entropy loss as the primary metric. To assess parameter efficiency across scales, we consider both 1-layer and 2-layer models at approximately 200K and 1M parameters. Exact hidden sizes, rank settings, and parameter counts for all model configurations are provided in Appendix~\ref{app:hyperparams}.

\paragraph{Larger-scale language modeling.}
We additionally evaluate on character-level text8 \cite{mahoney2011text8} with the standard 90M/5M/5M train/val/test split (vocabulary size 27), roughly $45\times$ larger than TinyShakespeare. Models are parameter-matched at approximately 193K parameters. For this statistical language-modeling setting we set the candidate-branch rank to $r_1 \approx H/2$ ($H=163$, $r_1=81$, $r_2=r_c=40$), reflecting the higher-rank demands of natural-language statistics relative to the compact algorithmic regime that is the focus of the paper. All text8 models are trained for 30 epochs with AdamW (learning rate $2\times10^{-3}$, $\beta=(0.9,0.99)$, weight decay $0.01$), batch size $128$, sequence length $200$, cosine decay, and gradient clipping at $1.0$; evaluation is reported in bits-per-character (bpc).

\paragraph{Formal-language task suite.}
To probe structured sequence reasoning beyond language modeling, we evaluate on a 7-task formal-language and algorithmic suite: \textsc{Parity}, \textsc{Counter}, \textsc{Lock}, \textsc{Dyck-2}, \textsc{Copy}, \textsc{Delay-2 Copy}, and \textsc{Noisy Counter}. These tasks cover complementary behaviors including long-range XOR-style recursion, modular counting, state locking, hierarchical bracket matching, exact sequence reproduction, delayed two-step dependency, and noise-robust accumulation. All formal-language data are generated online. Training sequences use lengths $T \in [8,64]$, while testing is divided into in-distribution (ID) lengths $[8,64]$, 2$\times$ out-of-distribution generalization (OOD1) lengths $[65,128]$, and 4$\times$ out-of-distribution generalization (OOD2) lengths $[129,256]$. Test sets are generated with fixed random seeds for reproducibility, and the length ranges are strictly separated between training and extrapolation evaluation. We report accuracy for each length regime. Unless otherwise noted, all models in the formal-language suite use hidden size $H=64$, yielding approximately 25K--34K parameters depending on the architecture.

\paragraph{Baselines and parameter matching.}
We compare baseVultra against standard PyTorch implementations of LSTM and GRU. For language modeling, baselines are parameter-matched as closely as possible within each model family and depth setting; exact hidden sizes and parameter counts are listed in Appendix~\ref{app:hyperparams}. For compute comparisons, we follow the metric defined in Sec.~\ref{sec:method}: dense recurrent matrix-vector operations under matched hidden size.

\paragraph{Training details.}
For character-level language modeling, all models are trained with AdamW \cite{loshchilov2019adamw} using learning rate $3\times 10^{-4}$, $\beta_1=0.9$, $\beta_2=0.95$, weight decay $0.1$, batch size $32$, sequence length $256$, and gradient clipping at $1.0$. We train for 100 epochs with a cosine decay schedule, using a 3-epoch warmup and minimum learning rate $3\times 10^{-5}$. Validation is performed every epoch on 30 batches. For formal-language tasks, we also use AdamW with learning rate $3\times 10^{-4}$, $(\beta_1,\beta_2)=(0.9,0.95)$, weight decay $0.1$, batch size $64$, cosine decay, 200 warmup steps, and gradient clipping at $1.0$, and train for 5{,}000 steps. Unless otherwise noted, we report mean $\pm$ standard deviation over three random seeds. Experiments were run on RTX 5090 and RTX 4060 Laptop GPUs.

\paragraph{Wall-clock and memory measurement protocol.}
Wall-clock time and memory are measured on parameter-matched models with batch size $32$ and sequence length $256$ on two GPUs (RTX 5090 and RTX 4060 Laptop). Training step time is the mean over a warmed-up run; peak memory is the maximum allocated during training. CUDA-graph capture on the 4060 uses static input shapes to remove per-step kernel-launch overhead without altering the computation.

\paragraph{Ethics and data.}
This work uses public-domain text and synthetically generated formal-language tasks. It does not involve human subjects, animal experiments, or personally identifiable information.

\subsection{Character-Level Language Modeling}
\label{sec:charlm}

\begin{table}[htbp]
\centering
\caption{Validation loss on TinyShakespeare (mean $\pm$ std over 3 seeds). Lower is better.}
\label{tab:charlm_main}
\begin{tabular}{lcccc}
\toprule
Model
& 1-layer $\sim$200K
& 1-layer $\sim$1M
& 2-layer $\sim$200K
& 2-layer $\sim$1M \\
\midrule
LSTM
& $1.6042 \pm 0.0071$
& $1.5174 \pm 0.0052$
& $1.6094 \pm 0.0100$
& $1.5038 \pm 0.0013$ \\
GRU
& $1.6159 \pm 0.0020$
& $1.5321 \pm 0.0079$
& $1.5797 \pm 0.0084$
& $\mathbf{1.4942 \pm 0.0022}$ \\
baseVultra
& $\mathbf{1.5939 \pm 0.0104}$
& $\mathbf{1.5172 \pm 0.0057}$
& $\mathbf{1.5739 \pm 0.0035}$
& $1.4994 \pm 0.0042$ \\
\bottomrule
\end{tabular}
\end{table}

We first evaluate baseVultra on TinyShakespeare to assess whether the proposed recurrent design improves sequence modeling quality under compact and medium-scale parameter budgets. Table~\ref{tab:charlm_main} reports validation loss for 1-layer and 2-layer recurrent models at approximately 200K and 1M parameters.

At the compact 1-layer 200K scale, baseVultra achieves the best validation loss among all compared recurrent baselines, reaching $1.5939 \pm 0.0104$ versus $1.6042 \pm 0.0071$ for a parameter-matched LSTM and $1.6159 \pm 0.0020$ for a parameter-matched GRU. This result is notable because baseVultra requires only about $54\%$ of GRU's dense recurrent matrix-vector computation under matched hidden size for the compact configuration used in our compute analysis. At the 1-layer 1M scale, baseVultra remains competitive, achieving $1.5172 \pm 0.0057$, effectively tied with LSTM and outperforming GRU.

A similar trend appears in the 2-layer 200K regime, where baseVultra again yields the best validation loss, outperforming both GRU and LSTM. At the 2-layer 1M scale, all three models become closer, with GRU achieving the lowest loss and baseVultra remaining competitive. Overall, these results support the central design hypothesis of this work: reallocating recurrent capacity toward candidate generation can improve nonlinear sequence modeling quality most clearly in compact regimes, while remaining competitive at larger scales.

\subsection{Larger-Scale Evaluation on text8}
\label{sec:text8}

\begin{table}[htbp]
\centering
\caption{Character-level text8 results (30 epochs, parameter-matched, single run). Bits-per-character, lower is better.}
\label{tab:text8}
\begin{tabular}{llrrr}
\toprule
Model & Configuration & Params & Val bpc & Test bpc \\
\midrule
baseVultra & $H=163$, $r_1=81$, $r_2=r_c=40$ & 192{,}829 & 1.797 & 1.854 \\
LSTM & $H=153$ & 192{,}627 & 1.783 & 1.837 \\
GRU & $H=175$ & 189{,}525 & 1.795 & 1.849 \\
\bottomrule
\end{tabular}
\end{table}

To test whether candidate-focused reallocation transfers beyond the compact algorithmic setting, we evaluate on text8, roughly $45\times$ larger than TinyShakespeare. Table~\ref{tab:text8} reports test bits-per-character for parameter-matched models. baseVultra is on par with GRU and $0.017$ bpc behind LSTM. All models shared the same 30-epoch compute budget, and validation loss was still decreasing for each, so the comparison is fair though not fully converged. Consistent with the regime-dependent thesis of this paper, the advantage of candidate-focused reallocation is most pronounced in compact settings, while full-gate cells remain competitive on large-scale statistical language modeling. We therefore state the operating boundary explicitly: we do not claim superiority for baseVultra at this scale.

Two diagnostic configurations corroborate the rank requirement on natural-language statistics. The paper's TinyShakespeare rank setting ($H=180$, $r_1=32$) reaches $1.876$ bpc, substantially worse than the $H/2$-rank configuration at the same parameter budget, confirming that natural-language statistics demand a higher-rank candidate branch than the compact algorithmic tasks. Relaxing the parameter budget slightly ($H=180$, $r_1=90$, $235$K parameters) reaches $1.830$ bpc, suggesting baseVultra remains competitive when the budget is allowed to grow.

\subsection{Formal-Language Tasks}
\label{sec:formal}

\begin{table}[htbp]
\centering
\small
\caption{Complete low-capacity formal-language results at hidden size $H=16$.
Accuracy on in-distribution (ID), $2\times$ OOD (OOD1),
and $4\times$ OOD (OOD2). Mean over 3 seeds; $\pm$ std shown for
non-saturated tasks. \textsc{Parity}, \textsc{Lock}, \textsc{Dyck-2}, and \textsc{Copy}
are saturated (1.000 at every seed for every model).}
\label{tab:formal_lowcap}
\begin{tabular}{llccc}
\toprule
Task & Model & ID & OOD1 & OOD2 \\
\midrule
\multirow{4}{*}{Counter}
& LSTM          & $0.866\pm0.008$ & $0.774\pm0.013$ & $0.691\pm0.018$ \\
& GRU           & $0.909\pm0.011$ & $0.819\pm0.016$ & $0.761\pm0.019$ \\
& baseVultra    & $\mathbf{0.999}$ & $\mathbf{0.996\pm0.001}$ & $\mathbf{0.991\pm0.003}$ \\
& bV-no\_h2     & $0.998\pm0.001$ & $0.989\pm0.007$ & $0.978\pm0.015$ \\
\midrule
\multirow{4}{*}{Delay-2 Copy}
& LSTM          & $0.889\pm0.004$ & $0.881\pm0.004$ & $0.879\pm0.004$ \\
& GRU           & $0.909\pm0.004$ & $0.904\pm0.004$ & $0.905\pm0.004$ \\
& baseVultra    & $\mathbf{0.977\pm0.006}$ & $\mathbf{0.976\pm0.007}$ & $\mathbf{0.975\pm0.008}$ \\
& bV-no\_h2     & $0.974\pm0.003$ & $0.972\pm0.002$ & $0.971\pm0.002$ \\
\midrule
\multirow{4}{*}{Noisy Counter}
& LSTM          & $0.396$ & $0.228$ & $0.168\pm0.001$ \\
& GRU           & $0.407\pm0.001$ & $0.233\pm0.001$ & $0.174$ \\
& baseVultra    & $0.411\pm0.001$ & $0.235\pm0.001$ & $0.174\pm0.001$ \\
& bV-no\_h2     & $0.398\pm0.001$ & $0.229\pm0.001$ & $0.170\pm0.001$ \\
\midrule
Parity      & all models & 1.000 & 1.000 & 1.000 \\
Lock        & all models & 1.000 & 1.000 & 1.000 \\
Dyck-2      & all models & 1.000 & 1.000 & 1.000 \\
Copy        & all models & 1.000 & 1.000 & 1.000 \\
\bottomrule
\end{tabular}
\end{table}

\begin{table}[htbp]
\centering
\small
\caption{Counter results at $H=32$ (mean over 3 seeds). All other tasks are
saturated (accuracy 1.000) or at chance (\textsc{Noisy Counter}) for every model
at this hidden size.}
\label{tab:formal_midcap}
\begin{tabular}{llccc}
\toprule
Model & ID & OOD1 & OOD2 \\
\midrule
LSTM          & $1.000$ & $1.000$ & $0.999\pm0.001$ \\
GRU           & $1.000$ & $1.000$ & $1.000$ \\
baseVultra    & $1.000$ & $0.998\pm0.001$ & $0.997$ \\
bV-no\_h2     & $1.000$ & $0.999$ & $0.998\pm0.001$ \\
\bottomrule
\end{tabular}
\end{table}

\begin{figure}[htbp]
\centering
\includegraphics[width=0.62\linewidth]{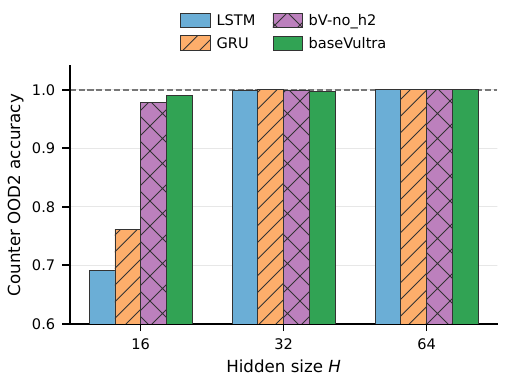}
\caption{%
\textbf{Counter OOD2 accuracy across hidden sizes.}
baseVultra shows the largest advantage at $H=16$, while all models
approach perfect accuracy as hidden size increases. The dashed line
marks perfect accuracy.}
\label{fig:counter_trend}
\end{figure}

We next evaluate baseVultra on formal-language and algorithmic sequence tasks as structural probes of recurrent behavior. In the standard $H=64$ regime, all three recurrent baselines---baseVultra, GRU, and LSTM---saturate nearly all tasks, reaching perfect or near-perfect accuracy across in-distribution and extrapolation settings. Rather than indicating the absence of structure, this result suggests that the benchmark becomes too easy at this hidden-state scale for all compared recurrent models.

To obtain a more discriminative probe, we study a low-capacity regime with smaller hidden states. This is the operating point of interest for this paper: candidate-focused capacity allocation is designed for exactly the setting where recurrent capacity is scarce.

Table~\ref{tab:formal_lowcap} reports the complete low-capacity results at $H=16$ for all seven tasks. On \textsc{Counter}, baseVultra substantially outperforms both GRU and LSTM, especially under extrapolation: at OOD2 it reaches $0.991$, compared with $0.761$ for GRU and $0.691$ for LSTM. A similar pattern appears on \textsc{Delay-2 Copy}, where baseVultra achieves $0.975$ on OOD2, compared with $0.905$ for GRU and $0.879$ for LSTM. Removing the second-order pathway (bV-no\_h2) reduces \textsc{Counter} OOD2 from $0.991$ to $0.978$---a modest, capacity-dependent contribution---and \textsc{Delay-2 Copy} OOD2 from $0.975$ to $0.971$. \textsc{Parity}, \textsc{Lock}, \textsc{Dyck-2}, and \textsc{Copy} are saturated at $H=16$ for all models, and \textsc{Noisy Counter} is at chance for all, providing ceiling and floor references for the suite.

At $H=32$ (Table~\ref{tab:formal_midcap}), only \textsc{Counter} retains any difference, and the gap has essentially closed (baseVultra $0.997$ vs.\ GRU $1.000$ OOD2); all other tasks are saturated or at chance for every model. At $H=64$, all seven tasks are saturated for every cell. As shown in Fig.~\ref{fig:counter_trend}, the \textsc{Counter} advantage disappears as hidden size increases, confirming that the benefit is capacity-dependent.

We also ran four additional second-order probes (\textsc{Delay-3}, \textsc{Delay-2} with interference, \textsc{Period-2}, and \textsc{Second-Difference}), comparing baseVultra against bV-no\_h2 under identical settings. Disabling $\mathbf{h}_{t-2}$ access stays within run-to-run noise on all four probes, corroborating the ablation result that the second-order pathway is a minor component.

Finally, we emphasize that OOD2 tests $4\times$ length extrapolation (lengths 129--256 vs.\ training 8--64). This is not an evaluation of long-document language modeling, which we did not perform and state as an explicit limitation of this work.

\subsection{Ablation Studies}
\label{sec:ablation}

\begin{table}[htbp]
\centering
\caption{Ablation results on 1-layer TinyShakespeare with $H=180$,
$r_1=32$, and $r_2=r_c=16$ (mean $\pm$ std over 3 seeds).
$\Delta$ denotes degradation relative to the full model.}
\label{tab:ablation}
\begin{tabular}{lcc}
\toprule
Variant & Validation Loss & $\Delta$ \\
\midrule
full             & $\mathbf{1.5967 \pm 0.0085}$ & -- \\
no\_h2           & $1.6001 \pm 0.0077$ & $+0.0034$ \\
no\_cache        & $1.6016 \pm 0.0070$ & $+0.0048$ \\
no primary LR    & $1.6040 \pm 0.0094$ & $+0.0073$ \\
no secondary LR  & $1.6103 \pm 0.0087$ & $+0.0136$ \\
no LR branches   & $1.6218 \pm 0.0104$ & $\mathbf{+0.0251}$ \\
\bottomrule
\end{tabular}
\end{table}

\begin{table}[htbp]
\centering
\caption{Minimal-core isolation on 1-layer TinyShakespeare ($H=180$,
$r_1=32$, $r_2=r_c=16$; mean $\pm$ std over 3 seeds). The Minimal variant
shares the identical architecture and total parameter count; every
auxiliary pathway is frozen at zero.}
\label{tab:minimal}
\begin{tabular}{lcc}
\toprule
Variant & Validation Loss & $\Delta$ \\
\midrule
Full baseVultra (rerun) & $\mathbf{1.5939 \pm 0.0066}$ & -- \\
Minimal (diagonal gates $+$ single dense candidate) & $1.6259 \pm 0.0030$ & $\mathbf{+0.0320}$ \\
\bottomrule
\end{tabular}
\end{table}

We next evaluate the contribution of individual design components on the 1-layer TinyShakespeare setting with $H=180$, $r_1=32$, and $r_2=r_c=16$. Table~\ref{tab:ablation} reports validation loss over three seeds. The full row is an independent rerun of the same configuration used for the ablation suite; all deltas are computed relative to this rerun. The largest degradation comes from removing both low-rank auxiliary branches: no LR branches increases validation loss from $1.5967 \pm 0.0085$ to $1.6218 \pm 0.0104$, a gap of $+0.0251$. This confirms that the multi-branch candidate-generation design is a primary source of performance gain.

Disabling individual branches reveals an asymmetric contribution. Removing the secondary low-rank branch ($U_2,V_2$) causes a larger degradation ($+0.0136$) than removing the primary low-rank branch ($U_1,V_1$, $+0.0073$), supporting the claim that the differentiated routing of the two low-rank paths leads to non-redundant functional roles rather than interchangeable corrections. Removing the EMA cache produces a smaller but consistent degradation ($+0.0048$), indicating that the cache provides a modest stabilizing benefit.

By contrast, removing the explicit second-order pathway (no\_h2) increases validation loss by only $+0.0034$, which is small relative to run-to-run variation on this language modeling setting. The targeted second-order probes in Sec.~\ref{sec:formal} confirm this: on \textsc{Delay-3}, \textsc{Delay-2} with interference, \textsc{Period-2}, and \textsc{Second-Difference}, disabling $\mathbf{h}_{t-2}$ stays within run-to-run noise. We therefore do not claim that second-order gating is a dominant factor. Instead, we interpret its role as capacity-dependent: the TinyShakespeare ablation suggests only a mild effect at this scale, whereas the low-capacity \textsc{Counter} results in Sec.~\ref{sec:formal} indicate a modest but measurable contribution ($0.991$ vs.\ $0.978$ OOD2) when hidden-state capacity is constrained.

To isolate the core contribution, we additionally construct a \emph{Minimal} variant with identical architecture and total parameter count, in which every auxiliary pathway is frozen at zero---the second-order terms ($\mathbf{w}_{rh2},\mathbf{w}_{zh2},\mathbf{w}_{ch2}$), both low-rank branches ($U_1V_1$ and $U_2V_2$), and the EMA cache ($\boldsymbol{\rho}\to0$, gain $\to0$)---leaving only the diagonal gates and the single dense candidate branch. Table~\ref{tab:minimal} shows the result: the Minimal core reaches $1.6259 \pm 0.0030$, a gap of $+0.032$ against the full model rerun ($1.5939 \pm 0.0066$), under identical parameters. The full rerun matches the main-table 1-layer 200K value ($1.5939$). Both the component-wise ablation and the minimal-core isolation point the same way: the differentiated multi-branch candidate design is the primary driver of baseVultra's gains, while the second-order term and the cache are minor refinements.

\subsection{Wall-Clock Time and Memory}
\label{sec:efficiency}

\begin{table}[htbp]
\centering
\caption{Training step time / peak memory on RTX 5090 (parameter-matched,
batch 32, sequence length 256).}
\label{tab:speed5090}
\begin{tabular}{lccc}
\toprule
Configuration & baseVultra & LSTM & GRU \\
\midrule
$H=420$ (1 layer)   & 24.4 ms / 591 MB  & 5.5 ms / 317 MB  & 5.9 ms / 375 MB \\
$H=1024$ (2 layers) & 58.7 ms / 1902 MB & 16.8 ms / 1252 MB & 13.9 ms / 1221 MB \\
$H=1400$ (3 layers) & 115 ms / 3479 MB  & 43.9 ms / 2415 MB & 37.4 ms / 2295 MB \\
\bottomrule
\end{tabular}
\end{table}

\begin{table}[htbp]
\centering
\caption{Training step time on RTX 4060 Laptop (parameter-matched,
batch 32, sequence length 256), standard execution and with CUDA-graph
capture.}
\label{tab:speed4060}
\begin{tabular}{lccc}
\toprule
Configuration & baseVultra (standard / graph) & LSTM \\
\midrule
$H=512$ (1 layer)   & 79.3 / 32.4 ms   & 20.6 ms \\
$H=1024$ (2 layers) & 195.1 / 114.6 ms & 111.7 ms \\
$H=1400$ (3 layers) & 368.4 / 270.5 ms & 351.3 ms \\
\bottomrule
\end{tabular}
\end{table}

We measure practical efficiency directly rather than relying on the theoretical FLOPs ratio. Tables~\ref{tab:speed5090} and~\ref{tab:speed4060} report training step time and peak memory for parameter-matched models on two GPUs. We are candid about the outcome: against the highly optimized cuDNN LSTM/GRU kernels, our reference implementation is $3$--$6\times$ slower and uses more memory, despite its lower dense-recurrent FLOPs and---at the larger benchmark configurations under matched hidden size---roughly $30\%$ fewer parameters than LSTM. The slowdown is dominated by per-step kernel-launch overhead rather than arithmetic cost. Two observations support this diagnosis. First, the gap narrows at larger scale on the slower 4060 but persists on the faster 5090, where launch overhead is relatively more expensive. Second, CUDA-graph capture---which removes per-step launch overhead without altering the computation---largely closes the gap: at $H=1400$, the step time drops from $368$ to $270$ ms, which is faster than the cuDNN LSTM step ($351$ ms). That a launch-overhead-free baseVultra can beat the cuDNN LSTM on wall-clock time at this configuration supports the diagnosis that the gap is launch cost rather than recurrent arithmetic. Inference latency follows the same kernel-launch-bound pattern.

We therefore scope the efficiency claim of this paper to dense recurrent matrix-vector FLOPs and parameter count under matched hidden size, and we do not claim a wall-clock speedup. Fused-kernel implementations and parallel-scan training are left to future work. As a correctness check, our custom kernel was verified against autograd, with maximum relative gradient error below $3\times10^{-5}$ for $H$ from 16 to 420 on two GPU architectures.

\section{Related Work}
\label{sec:related}

\paragraph{Gated recurrent architectures.}
LSTM and GRU remain the dominant reference points for nonlinear recurrent sequence modeling, primarily because they provide stable optimization and strong empirical performance with relatively simple gating mechanisms \cite{hochreiter1997lstm,cho2014gru}. More broadly, classical recurrent architectures such as Elman networks established the basic one-step recurrent paradigm on which later gated variants were built \cite{elman1990rnn}. Subsequent work has proposed richer recurrent dynamics, alternative gating rules, or more efficient state updates, but in practice these architectures have rarely displaced LSTM and GRU as standard nonlinear baselines. Recent renewed interest in efficient recurrent and recurrent-like sequence modeling includes structured state-space and linear-time architectures such as S4, Mamba, RWKV, xLSTM, RetNet, Griffin, and the broader re-examination of recurrent models for long sequences \cite{gu2022s4,gu2023mamba,peng2023rwkv,beck2024xlstm,sun2023retnet,de2024griffin,orvieto2023resurrecting}. Our work differs from this line in both emphasis and mechanism: rather than introducing additional dense recurrent transformations, baseVultra reallocates recurrent capacity toward candidate generation and replaces recurrent gate matrices with diagonal $O(H)$ interactions.

\paragraph{Higher-order recurrent models.}
Higher-order and multi-step recurrent dependencies have been explored in earlier recurrent literature, including second-order recurrent networks and models with richer dynamical recognizers \cite{giles1991second,pollack1991recursive}. Related work has also explored alternative recurrent dynamics for stability or expressive sequence modeling, including unitary and antisymmetric recurrent parameterizations \cite{arjovsky2016unitary,chang2019antisymmetric}. However, such ideas are rarely integrated into compact gated cells in a way that preserves practical compute efficiency. In baseVultra, second-order dependence is introduced through diagonal gate-history terms, allowing direct access to $\mathbf{h}_{t-2}$ at $O(H)$ cost without adding another dense recurrent matrix. Our claim is therefore not one of asymptotic expressiveness over first-order RNNs, but of improved state efficiency under constrained hidden size---and, as our ablations and probes show, the empirical contribution of this mechanism is marginal in most settings.

\paragraph{Formal-language evaluation.}
Formal-language and algorithmic sequence tasks have long been used as controlled probes of recurrent inductive bias, including counting, copying, locking, and hierarchical matching \cite{weiss2018practical,suzgun2019evaluating,deletang2023neural}. Such tasks are especially useful when standard language modeling benchmarks alone cannot reveal which structural behaviors a recurrent cell can or cannot support. Related analyses have also studied the limits of sequence architectures on formal languages and long-range structure, including both self-attention-based and recurrent models \cite{hahn2020theoretical,bhattamishra2020ability}. We adopt this perspective here, using formal tasks not as stand-alone benchmarks but as diagnostic probes of capacity-sensitive recurrent behavior. In particular, our low-capacity analysis highlights that differences between recurrent cells may disappear in saturated regimes and only become visible when hidden-state capacity is constrained.

\section{Conclusion}
\label{sec:conclusion}

We presented baseVultra, a compact nonlinear recurrent cell based on recurrent capacity reallocation. The cell concentrates the dense $H \times H$ recurrent matrix on candidate generation, replaces recurrent gate matrices with diagonal $O(H)$ interactions, and enriches the candidate with two low-rank auxiliary branches, an EMA cache, and learnable output blending. A lightweight second-order gating term provides direct access to $\mathbf{h}_{t-2}$ at $O(H)$ cost; we measured its contribution explicitly and found it marginal in most settings.

Empirically, baseVultra improves over parameter-matched LSTM and GRU baselines in compact character-level language modeling (TinyShakespeare), and is competitive but not better on the larger text8 benchmark, marking the operating boundary of the method. Component-wise ablation and the minimal-core baseline show that the differentiated multi-branch candidate design is the primary driver of these gains. On formal-language probes, the advantage is concentrated at constrained capacity (\textsc{Counter} OOD2 at $H=16$: $0.991$ vs.\ $0.761$ GRU and $0.691$ LSTM) and vanishes as hidden size grows.

The scope of these results is deliberately bounded. First, the benefit is concentrated in the compact regime; we do not claim superiority at large scale. Second, the efficiency claim concerns dense recurrent matrix-vector FLOPs and parameter count under matched hidden size; the reference implementation is $3$--$6\times$ slower than cuDNN kernels in wall-clock time, a gap dominated by per-step kernel-launch overhead that fused kernels and parallel-scan training may close in future work. Third, our formal-language probes test $4\times$ length extrapolation, not long-document language modeling, which we did not evaluate. Future work will pursue fused-kernel implementations, parallel-scan training, long-document evaluation, and additional large-scale benchmarks.

\paragraph{Use of generative AI.}
Generative AI tools were used to assist with language editing, LaTeX formatting, and figure-generation code. All content was reviewed and edited by the author(s), who take full responsibility for the submission.

\appendix

\section{Hyperparameters and Implementation Details}
\label{app:hyperparams}

\paragraph{Cell parameterization.}
Table~\ref{tab:basevultra_params} lists the learnable parameters of the baseVultra cell.
\begin{table}[htbp]
\centering
\small
\caption{Parameterization of the baseVultra cell.}
\label{tab:basevultra_params}
\begin{tabular}{lll}
\toprule
Parameter & Shape & Role \\
\midrule
$W_x$ & $4H \times d_{\mathrm{in}}$ & input projection weight \\
$\mathbf{b}_x$ & $4H$ & input projection bias \\
$W_{vh}$ & $H \times H$ & dense recurrent main branch \\
$\mathbf{w}_{rh}, \mathbf{w}_{zh}, \mathbf{w}_{ch}$ & $H$ & first-order diagonal gate weights \\
$\mathbf{w}_{rh2}, \mathbf{w}_{zh2}, \mathbf{w}_{ch2}$ & $H$ & second-order diagonal gate weights \\
$\mathbf{b}_r, \mathbf{b}_z, \mathbf{b}_c$ & $H$ & gate biases \\
$\boldsymbol{\beta}$ & $H$ & element-wise adaptive gain \\
$\Delta \mathbf{b}_v$ & $H$ & candidate offset \\
$V_1, U_1$ & $r_1 \times H,\; H \times r_1$ & primary low-rank branch \\
$V_2, U_2$ & $r_2 \times H,\; H \times r_2$ & secondary low-rank branch \\
$P_v, P_u$ & $r_c \times H,\; H \times r_c$ & cache projection matrices \\
$\boldsymbol{\rho}_{\mathrm{raw}}$ & $H$ & cache decay pre-activation \\
$\boldsymbol{\gamma}$ & $H$ & cache injection gain \\
$\boldsymbol{\ell}_1, \boldsymbol{\ell}_2$ & $H$ & output blending coefficients \\
\bottomrule
\end{tabular}
\end{table}

\paragraph{Model configurations.}
Table~\ref{tab:configs} lists all hidden sizes, rank settings,
and parameter counts used in character-level language modeling.

\begin{table}[htbp]
\centering
\small
\caption{Model configurations used in character-level language modeling.}
\label{tab:configs}
\begin{tabular}{lllr}
\toprule
Model & Scale & Configuration & Parameters \\
\midrule
baseVultra & 1L-200K & $H=180,\,r_1=32,\,r_2=r_c=16$ & 200,160 \\
baseVultra & 1L-1M   & $H=420,\,r_1=64,\,r_2=r_c=32$ & 1,024,800 \\
baseVultra & 2L-200K & $H=128,\,r_1=32,\,r_2=r_c=16$ & 209,792 \\
baseVultra & 2L-1M   & $H=300,\,r_1=64,\,r_2=r_c=32$ & 1,084,500 \\
\midrule
LSTM & 1L-200K & $H=153$ & 198,441 \\
LSTM & 1L-1M   & $H=349$ & 999,885 \\
LSTM & 2L-200K & $H=109$ & 198,925 \\
LSTM & 2L-1M   & $H=250$ & 1,020,250 \\
\midrule
GRU  & 1L-200K & $H=175$ & 196,175 \\
GRU  & 1L-1M   & $H=402$ & 998,166 \\
GRU  & 2L-200K & $H=126$ & 200,214 \\
GRU  & 2L-1M   & $H=289$ & 1,024,505 \\
\midrule
baseVultra & text8 & $H=163,\,r_1=81,\,r_2=r_c=40$ & 192,829 \\
LSTM & text8 & $H=153$ & 192,627 \\
GRU  & text8 & $H=175$ & 189,525 \\
\bottomrule
\end{tabular}
\end{table}

\begin{figure}[htbp]
\centering
\includegraphics[width=0.94\linewidth]{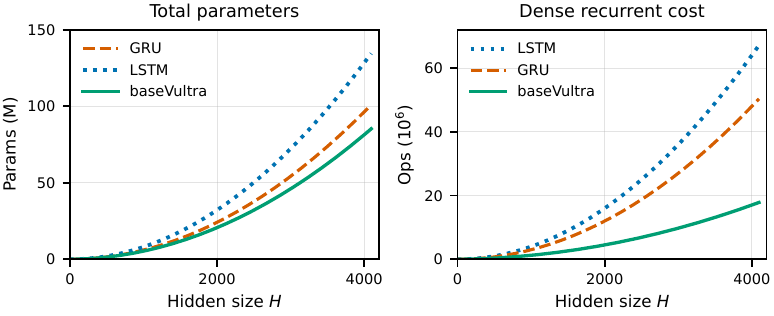}
\caption{%
Scaling behavior under matched hidden size.
\textbf{Left:} total parameters for 1-layer character-level models.
\textbf{Right:} dense recurrent matrix-vector operations per step.
With fixed ranks $(r_1=64,r_2=r_c=32)$, baseVultra uses $53.7\%$
of GRU dense recurrent cost at $H=420$ and $35.4\%$ at $H=4096$.}
\label{fig:scaling}
\end{figure}

\paragraph{Character-level language modeling.}
All models use AdamW with $\text{lr}=3\times10^{-4}$, $\beta_1=0.9$, $\beta_2=0.95$, weight decay $0.1$, batch size $32$, sequence length $256$, gradient clipping $1.0$, cosine decay schedule with 3-epoch linear warmup and minimum lr $3\times10^{-5}$, trained for 100 epochs. Validation uses 30 batches per epoch.

\paragraph{text8.}
All models use AdamW with $\text{lr}=2\times10^{-3}$, $\beta=(0.9,0.99)$, weight decay $0.01$, batch size $128$, sequence length $200$, gradient clipping $1.0$, and a cosine decay schedule, trained for 30 epochs on the standard 90M/5M/5M train/val/test split (vocabulary size 27). Evaluation is reported in bits-per-character.

\paragraph{Formal-language tasks.}
AdamW with $\text{lr}=3\times10^{-4}$, $(\beta_1,\beta_2)=(0.9,0.95)$, weight decay $0.1$, batch size $64$, 200-step warmup, cosine decay, gradient clipping $1.0$, trained for 5{,}000 steps.

\paragraph{Wall-clock and memory measurement.}
Parameter-matched models, batch size $32$, sequence length $256$, on RTX 5090 and RTX 4060 Laptop. Training step time is the mean over a warmed-up run; peak memory is the maximum allocated during training. CUDA-graph capture on the 4060 uses static input shapes to remove per-step kernel-launch overhead without altering the computation.

\paragraph{Minimal variant.}
The Minimal variant shares the identical architecture and total parameter count as the full model. Every auxiliary pathway is frozen at zero: the second-order terms ($\mathbf{w}_{rh2},\mathbf{w}_{zh2},\mathbf{w}_{ch2}$), both low-rank branches ($U_1V_1$ and $U_2V_2$), and the EMA cache ($\boldsymbol{\rho}\to0$, gain $\to0$). Only the diagonal gates and the single dense candidate branch remain active.

\paragraph{baseVultra initialization.}
All weight matrices use Xavier uniform initialization. Vector parameters and biases are initialized to zero. $\alpha$ is fixed to $\sqrt{e/2}\approx1.1658$ in all experiments and is not tuned per task. $\boldsymbol{\rho}_{\mathrm{raw}}$ is initialized to $0.0$ ($\boldsymbol{\rho}\approx0.5$). $W_{\mathrm{vh}}$ uses Xavier uniform. $\beta_{\max}=1.0$.

\paragraph{Assets and licenses.}
TinyShakespeare follows the publicly available character-level benchmark setup popularized by Karpathy \cite{karpathy2015tinyshakespeare}, based on public-domain Shakespeare text. text8 is the standard large text compression benchmark \cite{mahoney2011text8}. PyTorch baselines are implemented in PyTorch 2.x, which is distributed under the BSD 3-Clause License.

\bibliographystyle{unsrt}
\bibliography{refs}

\end{document}